\documentclass{ecai} 

\usepackage{latexsym}
\usepackage{amssymb}
\usepackage{amsmath}
\usepackage{amsthm}
\usepackage{booktabs}
\usepackage{enumitem}
\usepackage{graphicx}
\usepackage{color}

\newcommand{\BibTeX}{B\kern-.05em{\sc i\kern-.025em b}\kern-.08em\TeX}

\begin{document}


\begin{frontmatter}


\paperid{0} 


\title{Recording First-person Experiences to Build a New Type of Foundation Model}


\author[A]{\fnms{Dionis}~\snm{Barcari}}
\author[A]{\fnms{David}~\snm{Gamez}}
\author[B]{\fnms{Aliya}~\snm{Grig}}

\address[A]{Department of Computer Science, Middlesex University, London, UK}
\address[B]{Evolwe, 251 Little Falls Drive, Wilmington, New Castle County, Delaware 19808, United States}

\begin{abstract}
Foundation models have had a big impact in recent years and billions of dollars are being invested in them in the current AI boom. The more popular ones, such as Chat-GPT, are trained on large amounts of Internet data. However, it is becoming apparent that this data is likely to be exhausted soon, and technology companies are looking for new sources of data to train the next generation of foundation models.

Reinforcement learning, RAG, prompt engineering and cognitive modelling are often used to fine-tune and augment the behaviour of foundation models. These techniques have been used to replicate people, such as Caryn Marjorie. These chatbots are not based on people's actual emotional and physiological responses to their environment, so they are, at best, a surface-level approximation to the characters they are imitating.

To address these issues, we have developed a recording rig that captures what the wearer is seeing and hearing as well as their skin conductance (GSR), facial expression and brain state (14 channel EEG). AI algorithms are used to process this data into a rich picture of the environment and internal states of the subject. Foundation models trained on this data could replicate human behaviour much more accurately than the personality models that have been developed so far. This type of model has many potential applications, including recommendation, personal assistance, GAN systems, dating and recruitment.

This paper gives some background to this work and describes the recording rig and preliminary tests of its functionality. It then suggests how a new type of foundation model could be created from the data captured by the rig and outlines some applications. Data gathering and model training are expensive, so we are currently working on the launch of a start-up that could raise funds for the next stage of the project.
\end{abstract}

\end{frontmatter}

\section{Introduction}
Foundation models have had a big impact in recent years and billions of dollars are being invested in them in the current AI boom. The more popular ones, such as Llama, Chat-GPT and Dall-E, can generate plausible text and images in response to text and image prompts. These models are trained on large amounts of text and image data scraped from the Internet - often accessed through the Common Crawl repository. Researchers and technology companies are starting to realize that this data source could become exhausted soon, so they are looking for new sources of data to train the next generation of foundation models \cite{Villalobos2022,Seetharaman2024}.

After initial training the behaviour of foundation models can be fine-tuned through a combination of reinforcement learning, retrieval augmented generation (RAG) and prompt engineering. These techniques have been used to create fairly plausible imitations of individual people, such as Marilyn Monroe (https://www.soulmachines.com/) and Caryn Marjorie (https://caryn.ai). However, these chatbots are not based on people's actual physiological and emotional responses, so they are, at best, a surface level approximation to the characters that they are imitating.

To address these issues this paper describes a new type of recording rig that captures what the wearer is seeing and hearing and their emotional and physiological reactions to these stimuli. Data recorded by this rig could be used to build a new type of foundation model, a first person-foundation model (FPFM), that maps environmental stimuli to a person's emotional and physiological reactions, and maps a person's physiological and emotional states to their behaviour. First-person foundation models could provide much more realistic personality modelling, which could be used for personal assistants, generative adversarial networks (GAN), dating, recruitment and to develop a new type of recommendation engine. This recording rig could also help to address the predicted shortage of training data for the next generation of foundation models. 

The first part of this paper gives background information about foundation models, personality modelling with foundation models, the role of emotional and physiological states in decision-making, and similar hardware to our recording rig. Section~\ref{FirstPersonRecorder} describes the hardware and software of the first-person recorder that we have developed along with preliminary tests of its functionality. The final section discusses how data recorded by this rig could be used to train a FPFM, outlines some applications of FPFMs and suggests improvements that could be made to the recording rig in the future.

\section{Background}
\subsection{Foundation Models}
Foundation models often use a transformer architecture \cite{Vaswani2017} with hundreds of billions of parameters. They are trained on large amounts of data using considerable computer resources. For example, GPT-1 was trained on 5GB, GPT-2 on 40GB, GPT-3 on 45TB and MusicGen on 20,000 hours of audio \cite{Wu2023,Copet2024}. After training, foundation models are given a prompt, such as a text input, and they generate an output, such as text, code or images. Some examples of foundation models are given in Table~\ref{table:FoundationModels}.

\begin{table}[h]
    \caption{Examples of mappings carried out by some of the current foundation models.}
    \label{table:FoundationModels}
    \centering
    \renewcommand{\arraystretch}{1.5} 
    \begin{tabular}{p{2.3cm} p{1cm} p{3cm}} 
    \toprule
    Input & Output & Foundation Models \\
    \hline
    Text & Text & GPT 3.5 Claude 3 \\
    Text & Images & DALL-E, Stable Diffusion, GPT 4 \\
    Text & Code & CoPilot, CodeWhisperer \\
    Text & Music & MusicGen \\
    Text, images & Robot actions & RT-2 \\
    Text, images, DNA & Text & Med-PaLM M \\
    DNA & Cellular function & Geneformer, scGPT \\
    \bottomrule
    \end{tabular}
\end{table}

The behaviour of foundation models is often fine-tuned through a combination of reinforcement learning, retrieval augmented generation (RAG) and prompt engineering. In reinforcement learning the model generates multiple different outputs, which are ranked, typically by a human, and then it is trained to generate the desired output more frequently in the future. In retrieval augmented generation a vector database is created with a set of documents that the foundation model is required to use. When the user enters a query, a vector search identifies documents that are most relevant to the user's input and these are combined with the user's query in the final prompt that is sent to the model. Prompt engineering is a variety of techniques that are used to structure prompts to get desired outputs. These include intents, roles, chains of thought and output constraints \cite{Chen2023}.

\subsection{Personality Modelling with Foundation Models}
Foundation models can imitate the conversational styles of individual people. With simple prompt engineering, Chat-GPT can generate text in the style of personalities, such as Donald Trump, whose speeches and tweets were included in the original training data. More sophisticated models have been created by companies like Facebook (imitations of Snoop Dogg, Tom Brady and other celebrities) and UneeQ, who constructs digital avatars with different personalities (https://www.digitalhumans.com/). For the most part, these chatbots appear to be generated by a combination of reinforcement learning, RAG and prompt engineering. The only potential exception that we are aware of is Soul Machines (https://www.soulmachines.com/), whose website claims that their AI characters are based on a combination of LLMs and multimodal cognitive models. Details are lacking, but our best guess is that cognitive models are used to simulate the agent's state, which is combined with previous knowledge (retrieved using RAG) to build the LLM prompt. Similar work has been carried out by Park et al. \cite{Park2023} and Kirk et al. \cite{Kirk2023}.

\subsection{Emotions, Physiology and Decision-making}
It is becoming increasingly recognized that our emotions and physiological states play a central role in our decision-making and behaviour. Consider a person who is sitting in front of a beef burger in a restaurant. Suppose they are experiencing hunger, and they predict that eating the burger will cause them to experience pleasurable sensations and a feeling of satiety. In this case their current and predicted physiological state explain their action of eating the burger. On the other hand, suppose that the person is feeling sick or cares deeply about animals. In this case they will not eat the burger and this behaviour will, again, be explainable in terms of their current and predicted emotional and physiological states. This relationship between emotions, physiology and decision-making is nicely described by Damasio \cite{Damasio1994}, whose theory of somatic markers explains how we associate positive and negative emotions with objects in our environment. Goel's \cite{Goel2022} theory of tethered rationality is based on the idea that emotional and physiological states play a central role in the selection and initiation of behaviour. A close relationship between emotions and decision making has also been demonstrated in many psychological studies \cite{Lerner2015, George2016}.

Within AI research, the important role that emotion plays in cognition is coming to be more widely recognized. As Pessoa puts it "Emotion is not an 'add on' that endows a robot with 'feelings' (for instance, reporting or expressing its internal state). It allows the significance of percepts, plans, and actions to be an integral part of all its computations." \cite[p. 168]{Pessoa2019}. An AI model of a person that does not include their emotional and physiological reactions will, at best, approximate the surface level. It will not include the motivation behind their behaviours or the diversity of people's behaviours (the different burger-eating outcomes in our example).

\subsection{Lifelogging and Google StreetView}
A variety of life logging devices have been created to store continuous records of people's lives. For example, the Narrative Clip, worn in front of the user, takes a picture every 30 seconds, and many phone apps have been developed to track different aspects of users' lives. As far as we are aware, no-one has developed a life logging device that captures both the environment and the emotional and physiological states of the wearer. Our recorder is also similar to Google StreetView, which uses cameras mounted on cars and people to capture exterior and interior environments.

\section{First-person Recorder}\label{FirstPersonRecorder}
Our hypothesis is that a foundation model that is trained from scratch on the stimuli and emotional and physiological states of a person will replicate human behaviour more effectively than surface-level approximations built with LLMs, RAG, cognitive models and prompt engineering. First-person recorders could also help to plug the estimated shortfall of data that is required to train the next generation of foundation models \cite{Villalobos2022,Seetharaman2024}. To test this hypothesis, we have built a recording rig that captures high quality data from the wearer and stores it in a format that is suitable for training a foundation model.

\subsection{Hardware and Software}
\begin{figure}[h]
    \centering
    \includegraphics[width=0.75\linewidth]{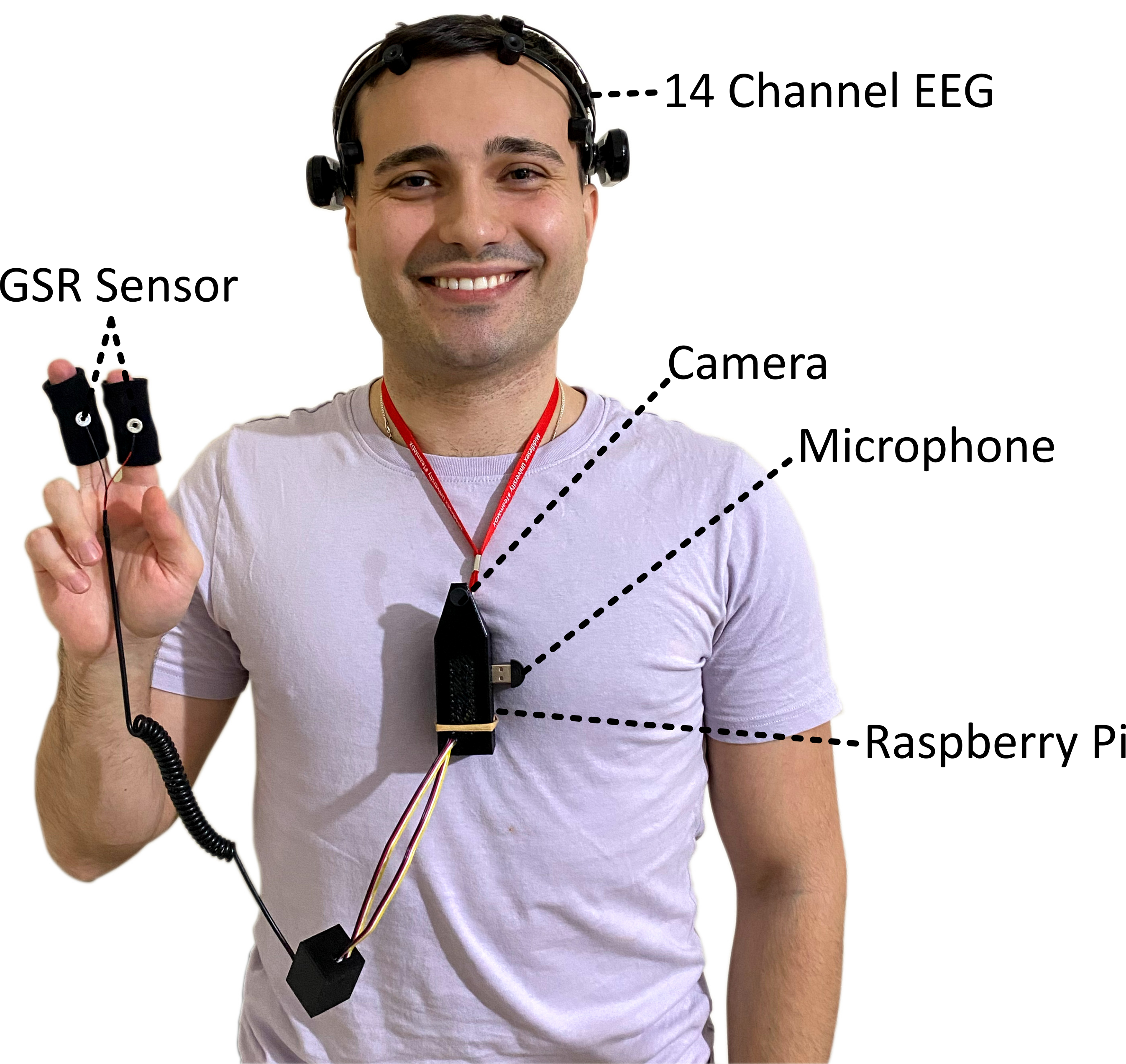}
    \caption{First-person recorder.}
    \label{fig:FirstPersonRecorder}
\end{figure}

\begin{figure}[h]
    \centering
    \includegraphics[width=0.9\linewidth]{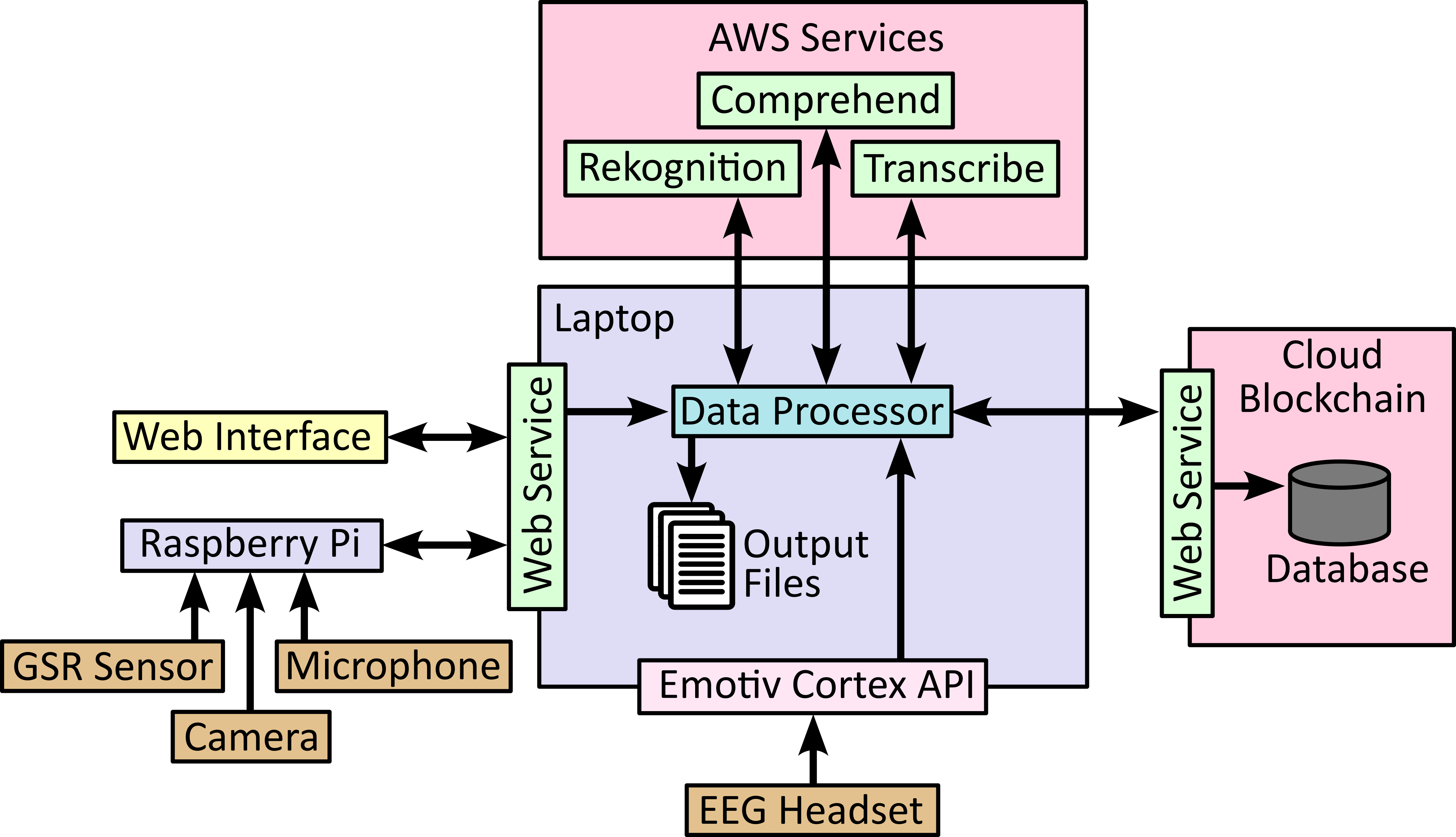}
    \caption{Architecture. The Raspberry Pi worn around the user's neck collects data from the camera, microphone and GSR sensor and sends it to a web service running on a laptop worn on the user's back. The Emotiv Epoc X EEG headset sends data to the laptop using the Emotiv Cortex API. A speaker attached to the Raspberry Pi emits a tone at random intervals for descriptive experience sampling (DES) \cite{Hurlburt2006}. A data processing program running on the laptop collects the data and uses Amazon Web Services (AWS) to identify text, sentiment and image labels. When the data for the file is complete, a hash of the file contents is sent to another cloud web service, which stores the original hash and sends back a new hash that combines the original hash with a random number. A web interface is provided to configure the recorder and play back recorded data.}
    \label{fig:Architecture}
\end{figure}

\begin{figure}[h]
    \centering
    \includegraphics[width=0.9\linewidth]{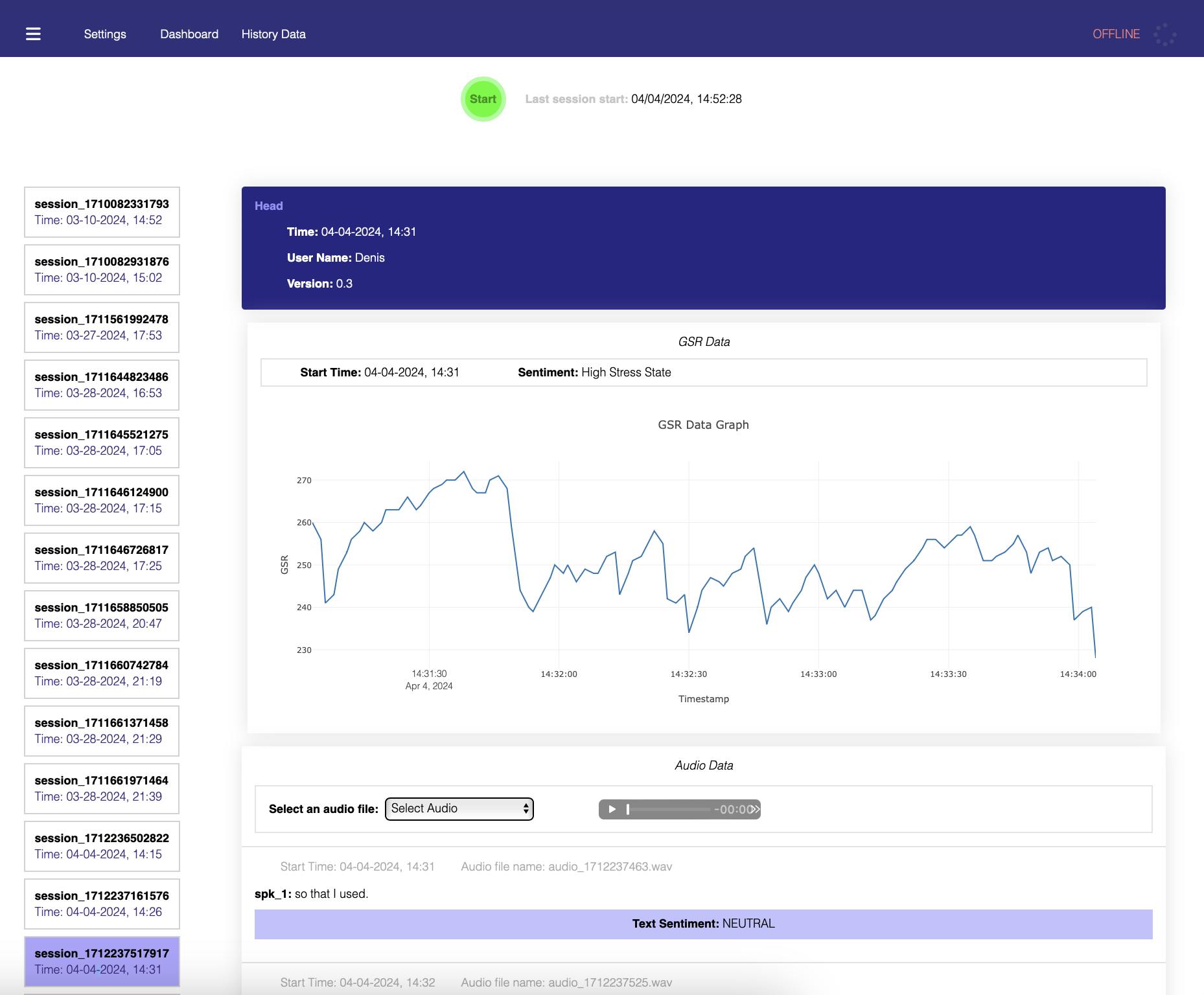}
    \caption{Web interface for recorder.}
    \label{fig:WebInterface}
\end{figure}

The recorder is based on a Raspberry Pi, worn around the user's neck, which is connected to a camera, microphone, GSR sensor and speaker. Data recorded by the Raspberry Pi is sent to a web service running on a laptop carried by the user, which has a WebSocket connection to the Epoc X EEG headset. Cloud services, such as AWS Rekognition, and the Emotiv Cortex API\footnote{https://emotiv.gitbook.io/cortex-api}, are used to analyse the raw data for higher level properties, such as text contents, sentiment, cognition, facial expression, and object labels. A website hosted on the laptop enables the recorder to be configured and supports playback of recorded data (see Figure~\ref{fig:WebInterface}). To reduce fraud a blockchain architecture is implemented that sends a hash of the data to the cloud and receives a hash of the data plus a random number known only to the cloud service, which is added to the next file in the sequence. To ensure the privacy of other people, all faces are automatically blurred during the recording process. The data is stored in JSON files; schema definitions for these files are available on the project website. Version 1.0 of the recorder is shown in Figure~\ref{fig:FirstPersonRecorder}. The architecture is illustrated in Figure~\ref{fig:Architecture} and the data stored by the recorder is summarized in Table~\ref{table:RecorderData}. 

\begin{table}[h]
    \caption{Data recorded by the rig. Schemas for the output files and an example recording are available at the project's GitHub page. Sample rates, such as image frequency, are configurable through the web interface. The data rates are based on the test recording on the project's GitHub repository, which contains one image per second. The data rates can vary, depending on the behaviour and environment of the user.}
    \label{table:RecorderData}
    \centering
    \renewcommand{\arraystretch}{1.5} 
    \begin{tabular}{p{1.5cm} p{1.5cm} p{3cm} p{1cm}}  
    \toprule
    Data & Source & Description & Rate (Kbps)\\
    \hline
    EEG & Emotiv Epoc X headset & Raw EEG data from 14 channels & 30 \\
    Audio & USB microphone & Audio from microphone mounted on front of user. & 20\break (mp3) \\
    Images & ZeroCam & Pictures from camera mounted in front of user. & 600\break (jpg) \\
    GSR & Grove GSR sensor & Galvanic skin response (GSR) of user. & 0.01\break (1 Hz) \\
    EEG band power & Emotiv\break Cortex API & EEG power in the theta, alpha, beta L, beta H, and gamma bands. & 8 \\
    Facial expression & Emotiv\break Cortex API & Eye action and expression on upper and lower face. & 4 \\
    Cognition & Emotiv\break Cortex API & Cognitive states, including engagement, excitement, stress, relaxation, interest and focus. & 0.02 \\
    Audio text & AWS\break Transcribe & Recorded audio is converted to text using AWS Transcribe. Speech of wearer is automatically separated out from other people's speech using an audio sample generated by the user. & 0.003 \\
    Speech sentiment & AWS\break Comprehend & Text generated by user is analysed for sentiment (positive, negative, mixed, and neutral) using AWS Comprehend. & 0.002 \\
    DES & User & Descriptive Experience Sampling (DES) is a technique in which a person describes the contents of their consciousness when prompted by a tone \cite{Hurlburt2006}. Key phrases, such as “Start Ziggy” and “End Ziggy”, identify the start and end of a DES report. & 0.001\break (per report) \\
    Image text & AWS\break Rekognition & Text in recorded images is identified using AWS Rekognition. & 0.001 \\	
    Image labels & AWS\break Rekognition & Labels for objects in recorded images are generated using AWS Rekognition. & 2 \\
    \bottomrule
    \end{tabular}
\end{table}

Code, 3D print files, JSON schemas for the data files, user manual and sample recordings are available at the project's GitHub repository: https://github.com/ancara22/Data-Recording-Rig.

\subsection{Preliminary Experiments}
Initial tests were carried out in a live environment on campus at Middlesex University, London. These tests established that the hardware could capture, process and store the data in real time. A sample of data recorded during these tests can be downloaded from the project's GitHub repository.

\begin{figure}[h]
    \centering
    \includegraphics[width=\linewidth]{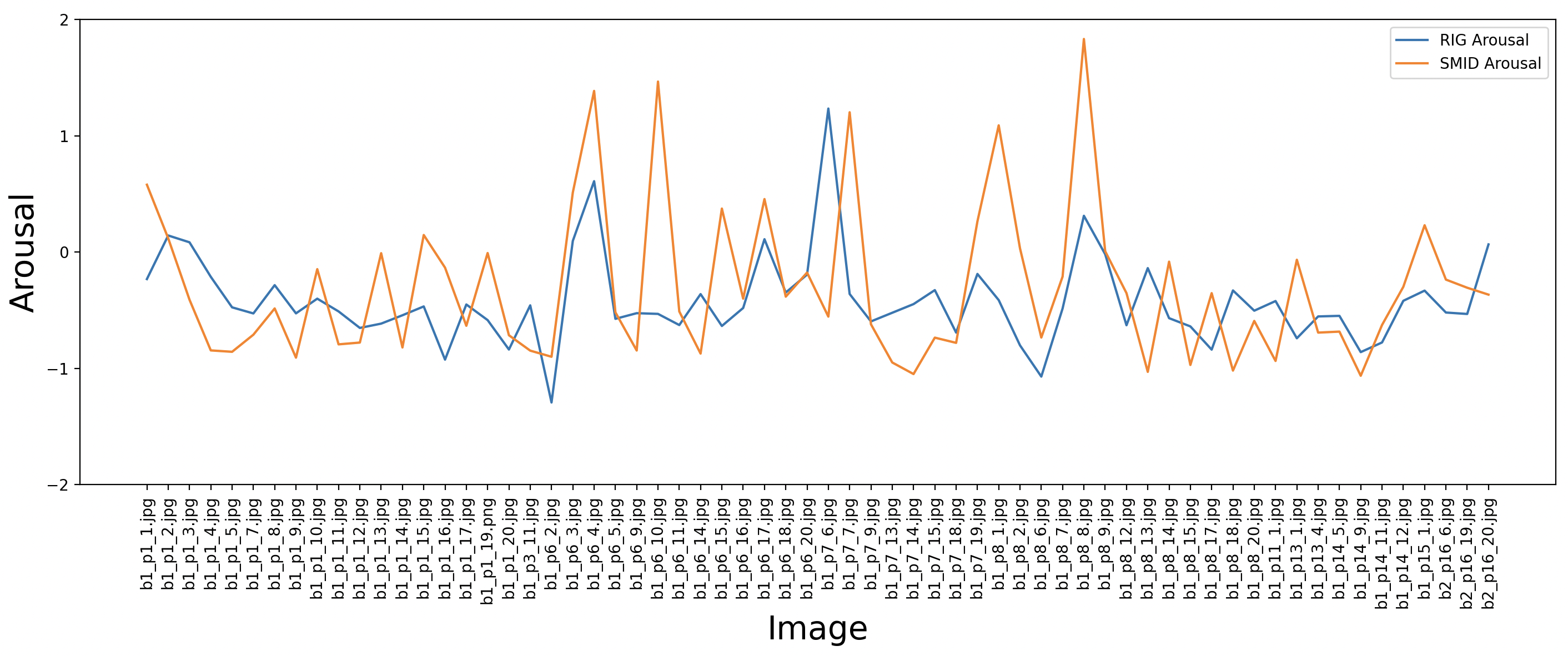}
    \caption{Mean arousal levels in experiments by Crone et al. \cite{Crone2018} and arousal levels recorded from a subject while viewing a sample of images from the SMID data set.}
    \label{fig:ArousalGraph}
\end{figure}

A second experiment was carried using images from the Socio-Moral Image Database (SMID), which were selected by Crone et al. \cite{Crone2018} for their ability to elicit specific emotions in the viewer. Crone et al. also measured the valence and arousal that were elicited by these images in 2,716 participants. To compare the data captured by our recorder with the results from Crone et al.'s experiments, a proxy for arousal was calculated from the combination of excitement and stress (see Table~\ref{table:RecorderData}) scaled to a value between -2.5 and 2.5. A selection of SMID images were then presented for 20 seconds each to a person wearing the recorder. The results are plotted in Figure~\ref{fig:ArousalGraph}.

The results presented in Figure~\ref{fig:ArousalGraph} are only given to illustrate how data recorded by the rig could be connected to the emotional and physiological states of a subject. Much more work needs to be done to combine the different data streams into a single picture of the subject's emotional and physiological state (see Section~\ref{NextSteps}).

\section{Discussion}
\subsection{First-person Foundation Models (FPFMs)}
The primary aim of this work is to record data from individuals and train a first-person foundation model that can carry out the following mappings:

\begin{itemize}

    \item \textit{Image/audio/text -> Emotional/physiological state}. How people's emotional and physiological state changes in response to different stimuli. For example, I feel sad when I look at a picture of my deceased grandmother.

    \item \textit{Emotional/physiological state -> External behaviour}. What people do or say when they are feeling a particular way. For example, I engage in food-seeking behaviour when I am hungry.

    \item \textit{Image/audio/text + Emotional/physiological state -> External behaviour}. What people do or say when they are feeling a certain way and experience a particular stimuli. For example, I eat a burger when I perceive a burger in front of me and I am hungry.

\end{itemize}

The data in Table~\ref{table:RecorderData} shows that the recorder can capture \textasciitilde40 GB of data (images, audio and text) in a 16 hour day. If the images, audio, raw EEG and raw GSR are excluded, this figure drops to \textasciitilde1 GB per 16 hour day. Based on these figures, Table~\ref{table:RecordingTimes} gives some rough estimates of how long it would take to store enough data to train foundation models on the scale of GPT-1, GPT-2 and GPT-3.

\begin{table}[h]
    \caption{Time required to record enough full data or text data (without raw GSR or EEG) to train GPT-1, GPT-2 and GPT-3. The recording times are based on 16 hour days.}
    \label{table:RecordingTimes}
    \centering
    \renewcommand{\arraystretch}{1.5} 
    \begin{tabular}{p{1cm} p{1.2cm} p{2cm} p{2.5cm}}  
    \toprule
    Model & Training Data (GB) & Recording Time\break (days, full) & Recording time\break (days, selected text)\\
    \hline
    GPT-1 & 5 & 0.14 & 6.5 \\
    GPT-2 & 40 & 1.1 & 52 \\
    GPT-3 & 46080 & 1300 & 60000 \\
    \bottomrule
    \end{tabular}
\end{table}

According to Table~\ref{table:RecordingTimes}, a purely text based version of GPT-2 could be created from \textasciitilde50 days of data from a single individual. Larger models are likely to require data recorded from several individuals.

To obtain this data one or more people could be paid to wear the recorder for an extended period of time. This could be accomplished through companies, such as Surge AI (https://www.surgehq.ai/), who recruit people to generate training data and fine tune AI models. Another option would be to release a cheap version of the recorder and launch a marketplace where people could be paid for their recordings. People could also be motivated to use the recorder in exchange for benefits, such as life logging, personal assistance, enhanced recommendation, or generation of media content based on their lives.\footnote{This was dramatized in the Joan is Awful episode of \textit{Black Mirror}.} This would be similar to the way in which big technology companies give us free and useful services, such as email, calendars and social networking, in exchange for access to our personal data.

When enough data has been collected, it could be used to train one or more first-person foundation models. Data gathering and model training are expensive, so we are currently working on the launch of a start-up that could raise funds for the next stage of the project.

\subsection{Applications of First-person Foundation Models}\label{Applications}
FPFMs have many potential applications:

\begin{itemize}

    \item \textit{Recommendation}. Modern recommendation systems are often based on keywords that are associated with films, music, or products. The keywords of previous products that the customer has consumed are mapped in a high dimensional space. Clustering and other techniques are then used to identify similar products to the ones that the customer has already consumed. FPFMs could be used to create a new form of recommendation engine that measures the customer's actual preferences for each product and recommends the ones with the highest net valence to the customer. For example, a FPFM based on \textit{my} emotional reactions could watch every single film and TV series on Netflix and recommend the ones that generate the most positive emotional states. 

    \item \textit{Focus groups}. FPFMs based on target audiences could be used to evaluate films, products,  political policies, etc. prior to their release.

    \item \textit{Dialog in novels and scripts}. Current foundation models, such as GPT-4, are already being used to generate novels and scripts. FPFMs could model characters more effectively, leading to more realistic dialog. It would also be possible to use FPFMs based on recordings of individual actors to generate scripts tailored to these actors. For example, a FPFM of Tom Cruise could be used to write the script for the next \textit{Mission Impossible} movie.

    \item \textit{Personal assistants}. A FPFM that understands the users preferences could book holidays, restaurants, etc. based on their preferences.

    \item \textit{GAN systems}. A FPFM could be used as the discriminator in a generative adversarial network, providing feedback about whether text, music, images, etc. generated by a foundation model are likely to produce positive emotional responses in a specific consumer. This could be a powerful way of improving the output quality of foundation models without human feedback.  

    \item \textit{Dating and recruitment}. Volar (https://www.volardating.com) uses the interactions between chatbots based on two people to reduce the awkwardness of first dates. Interactions between FPFMs of prospective partners would be a much more accurate way of evaluating the suitability of a match\footnote{This was dramatized in the Hang the DJ episode of \textit{Black Mirror}.}. A similar method could be used to evaluate whether job candidates are compatible with a company's current team.

\end{itemize}

Other applications include bereavement support, assisting dementia patients with a model of their former selves, patient models for psychologist training and phobia treatment.

\subsection{Next Steps}\label{NextSteps}
The backend of Version 1.0 of the recorder was implemented in a laptop carried in a backpack to reduce development costs. In the future this backend could be migrated to the cloud. While the Raspberry Pi could be replaced with a smartphone app, the battery drain is likely to be prohibitively high and people's frequent use of their phones would compromise the quality of the recordings. Something like the Humane AI pin (https://humane.com) would be a more suitable upgrade. Future versions of the recorder could also pull data from physiological trackers, such as the Fitbit.

Further work is required to integrate the recorded signals into a single picture of the emotional and physiological state of the wearer. This could be done through a machine learning approach, in which the DES data is used as the labels and a deep network is trained to automatically generate the labels from the recorded data. Since people's emotional and physiological reactions vary widely, it is likely to be necessary to calibrate the recorder for each user. 

The camera mounted on the user's chest captures their field of view, but not what they are actually looking at. One solution would be to use eye-tracking technology - for example, the Tobii Pro\footnote{See: https://www.tobii.com/.} - to record what the wearer is seeing. High quality eye-tracking hardware is expensive, so this could be approximated by increasing the resolution of the camera and using an attention model \cite{IttiKoch2001} to make a best guess about the contents of the wearer's vision. The attention model could be combined with top-down knowledge of which objects are likely to be salient to the user. For example, if the image contained a spider and the user had strong negative reactions to spiders, then it is likely that their eyes would saccade to the spider.

The first-person recorder captures everything that the wearer is exposed to, including copyrighted books, music, and films. To reduce this problem, a FPFM based on a single individual could be used \textit{privately} by that individual (for recommendation, etc. - see Section~\ref{Applications}) without being made available to third parties. This would also ensure the privacy of the user's data. Another option would be to use GPS to switch the recorder off automatically in situations in which copyright is likely to be an issue, such as concerts and cinemas. Recordings could also be automatically screened for copyrighted content. The elimination of copyrighted content from the model would reduce the power of the FPFM as a recommendation engine, since it would no longer be storing the user's reactions to books, films, and music. 

The automatic blurring of faces that was introduced to protect privacy would prevent a FPFM trained on the data from learning the strong emotional reactions that we have to familiar faces, such as friends, family, colleagues and celebrities. To address this issue people could be asked to give their consent to have their faces recorded by the device. This could be at an individual level (I give my consent for my wife to record my face on her device) or globally (celebrities could give their consent to be recorded by anyone). Face recognition could then be used to record the faces of consenting people in an unblurred state.

\section{Conclusion}
The central role that emotions and physiology play in the selection and initiation of behaviour is widely acknowledged in neuroscience and psychology, and it is starting to be recognised in artificial intelligence research. If this interpretation of the importance of emotion is correct, foundation models that are trained on text and image data from the Internet will only be able to create surface-level approximations to the behaviour of individual people. Technology companies are also starting to understand the limits of data scraped from the Internet. Many are now searching for new sources of data that could be used to train the next generation of foundation models.

To address these issues, we have developed a recording rig that stores what the wearer is seeing and hearing as well as their emotional and physiological reactions to their environment. A first-person foundation model trained on this data would be able to model human minds much more realistically than existing systems. There are many other exciting applications of FPFMs, including recommendation, focus groups, dialog writing, GAN systems, dating and recruitment. We are currently working on the launch of a start-up that could raise the funds to record a substantial amount of first-person data and train a FPFM model.



\begin{ack}
We would like to express our thanks to Evolwe (https://evolwe.ai) for financially supporting this work. We are particularly grateful to Alexander Morozov and Anastasia Rizzo from Evolwe, for their helpful feedback throughout the project. This research was approved by the ethics committee at Middlesex University, London.
\end{ack}



\bibliography{mybibfile}

\end{document}